\documentclass[fleqn,10pt]{wlscirep}
\usepackage[utf8]{inputenc}
\usepackage[T1]{fontenc}
\usepackage{lineno}
\usepackage{graphicx}
\usepackage{diagbox}
\usepackage{multirow}
\usepackage{CJKutf8}
\usepackage{pinyin}
\usepackage{xcolor}
\usepackage{subfigure}
\usepackage[ruled,linesnumbered]{algorithm2e}

\newcommand{\dataset}{EVOBC}
\newcommand{\zh}[1]{\begin{CJK*}{UTF8}{gbsn}#1\end{CJK*}}

\title{An open dataset for the evolution of oracle bone characters: EVOBC}

\author[1,$\dag$]{Haisu Guan}
\author[1,$\dag$]{Jinpeng Wan}
\author[1]{Pengjie Wang}
\author[1]{Kaile Zhang}
\author[1]{Zhebin Kuang}
\author[3]{Xinyu Wang}
\author[4]{Shengwei Han}
\author[4]{Yongge Liu}
\author[1]{Xiang Bai}
\author[2]{Lianwen Jin}
\author[1,*]{Yuliang Liu}
\affil[1]{Huazhong University of Science and Technology, Wuhan, 430074, China}
\affil[2]{South China University of Technology, Guangzhou, 510641, China}
\affil[3]{The University of Adelaide, SA, 5005, Australia}
\affil[4]{Anyang Normal University, Anyang, 455000, China}
\affil[*]{Corresponding author(s): Yuliang Liu (ylliu@hust.edu.cn)}

\affil[$\dag$]{These authors contributed equally to this work}

\begin{abstract}

The earliest extant Chinese characters originate from oracle bone inscriptions, which are closely related to other East Asian languages. These inscriptions hold immense value for anthropology and archaeology. However, deciphering oracle bone script remains a formidable challenge, with only approximately 1,600 of the over 4,500 extant characters elucidated to date. Further scholarly investigation is required to comprehensively understand this ancient writing system. Artificial Intelligence technology is a promising avenue for deciphering oracle bone characters, particularly concerning their evolution. However, one of the challenges is the lack of datasets mapping the evolution of these characters over time. In this study, we systematically collected ancient characters from authoritative texts and websites spanning six historical stages: Oracle Bone Characters - \textbf{OBC} (15th century B.C.), Bronze Inscriptions - \textbf{BI} (13th to 221 B.C.), Seal Script - \textbf{SS} (11th to 8th centuries B.C.), Spring and Autumn period Characters - \textbf{SAC} (770 to 476 B.C.), Warring States period Characters - \textbf{WSC} (475 B.C. to 221 B.C.), and Clerical Script - \textbf{CS} (221 B.C. to 220 A.D.). Subsequently, we constructed an extensive dataset, namely EVolution Oracle Bone Characters (EVOBC), consisting of 229,170 images representing 13,714 distinct character categories.
We conducted validation and simulated deciphering on the constructed dataset, and the results demonstrate its high efficacy in aiding the study of oracle bone script. This openly accessible dataset aims to digitalize ancient Chinese scripts across multiple eras, facilitating the decipherment of oracle bone script by examining the evolution of glyph forms.
\end{abstract} 

\begin{document}
\flushbottom
\maketitle

\thispagestyle{empty}

\section*{Background \& Summary}

\begin{figure}[ht!]
\centering
\includegraphics[width=0.9\linewidth]{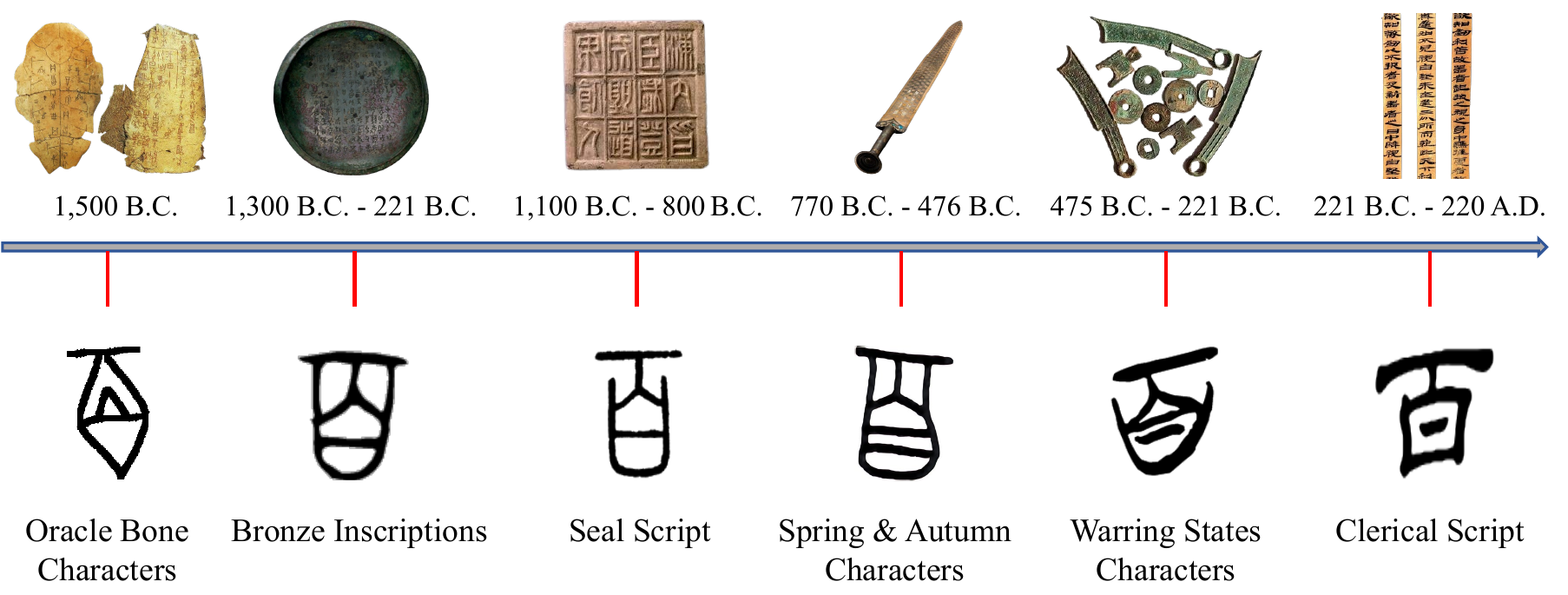}
\caption{The evolutionary path of the Chinese character ``\zh{百}'' (\bai3).}
\label{fig:intro-development}
\end{figure}

Written language stands as one of the pivotal symbols of humanity's ascent into civilized society. From the hieroglyphs of ancient Egyptians to the cuneiform script of the Sumerians in the Mesopotamian region, from Mayan glyphs across the American continent to the Oracle Bone Script in ancient China's Shang Dynasty, each script has meticulously recorded the dawn and legacy of splendid human cultures. For thousands of years, these scripts have been crafted and utilized by humans, recording everything from the Code of Hammurabi in cuneiform to the narratives of the Bible in Hebrew, showcasing their integral role in shaping history and civilization.

Among these ancient scripts, the Oracle Bone Script of China holds a unique position. Originating over 3,000 years ago during the Shang Dynasty, it is distinguished not only as the ancestor of modern Chinese characters but also as a vivid testament to the unbroken evolution of written language. The Oracle Bone Script embodies a rare continuity, offering a direct lineage that can be traced from its earliest forms to the contemporary Chinese character used today. For example, Figure~\ref{fig:intro-development} illustrates the evolution of the modern standard Chinese character \zh{百}(\bai3) from its oracle bone script form dating back to the Shang Dynasty 1,500 B.C. This evolutionary path is not merely of academic interest; it provides an uninterrupted historical thread, linking modern Chinese with its ancient roots in a way that few other scripts can claim. Its extensive use in a wide range of activities, from divination to recording aspects of daily life, has left a rich legacy, making it an invaluable resource for historians and linguists.

Since their discovery in 1899, Oracle Bone inscriptions have captivated numerous scholars, sparking intense research efforts. The National Museum of Chinese Writing has even offered up to 100,000 RMB per character for successfully deciphering uninterpreted Oracle Bone characters. Despite this, of the approximately 4,500 individual Oracle Bone characters discovered to date, two-thirds remain undeciphered, their meaning still shrouded in mystery~\cite{ZGXE201923047}. Given the recent advancements in Artificial Intelligence (AI), especially the remarkable success of Optical Character Recognition (OCR) techniques in modern text recognition, a pertinent question arises: \textit{Could these AI methods be effectively employed in deciphering Oracle Bone script?} However, a major hurdle remains: AI models generally need extensive data for training and validation, and as of now, there's no comprehensive dataset to support AI-assisted deciphering of Oracle Bone inscriptions. Therefore, this paper aims to construct such a dataset to support future research in this area.

The evolution of written characters did not happen overnight. For instance, Wang et al.~\cite{wang2022study} discovered a notable connection between Oracle Bone Character and Bronze Inscriptions, suggesting the potential to decipher Oracle Bone Characters using texts from other periods that have already been understood. Inspired by this, we constructed the EVolution of the Oracle Bone Characters (\dataset{}) dataset. This dataset categorizes ancient Chinese characters into six key periods, as illustrated in Figure~\ref{fig:intro-development}: Oracle Bone Characters (OBC), Bronze Inscriptions (BI), Seal Script (SS), Spring \& Autumn period Characters (SAC), Warring States Characters (WSC), and Clerical Script (CS). Among these, CS is closest to modern standard Chinese characters, while OBC originates from the most distant past. The \dataset{} dataset collates the representations of the same Chinese character across up to six different periods, forming a clear evolutionary trajectory for each character.

To build such a dataset, it is necessary to gather a vast array of images and labels from various sources. Manually collecting these scans of ancient texts demands significant effort and cost. Therefore, we have devised an automated process to extract both images and corresponding labels from diverse sources, including books and online databases. The final product, \dataset{}, includes a total of 229,170 images in 13,714 different categories. Specifically, 90,882 images from 8,376 categories were extracted from books, and 138,288 images from 10,207 categories were sourced from online repositories. We hope that this dataset will contribute to future computer-assisted research in deciphering Oracle Bone Characters.

\section*{Methods}

\subsection*{Data Source}

\begin{table}[h]
\centering
\small
\begin{tabular}{cccccc}
Source & Type & Period & Era & \#Categories & \#Images \\\hline
YinQiWenYuan~\cite{YinQiWenYuan} & Web & Shang Dynasty & 1617 B.C. - 1046 B.C. & 1,119 & 1,633 \\
GuoXueDaShi~\cite{GuoXueDaShi} & Web & All Period & / & 10,158 & 106,010 \\ 
Oracle Bone Character Compilation~\cite{jgw} & Book & Shang Dynasty & 1617 B.C. - 1046 B.C. & 1,202 & 17,600 \\
Compilation of Western Zhou Bronze Inscription~\cite{jinwen} & Book & Western Zhou & 1046 B.C. 
 - 771 B.C. & 2,040 & 21,681 \\
Spring and Autumn Script Glyph Table~\cite{chunqiu} & Book & Spring\&Autumn & 770 B.C. - 476 B.C. & 1,905 & 9,131 \\
Table of Glyphs for Warring States~\cite{zhanguo} & Book & Warring States & 475 B.C. - 221 B.C. & 6,658 & 32,794 \\\hline
\end{tabular}
\caption{Data source of the \dataset.}
\label{table:data-source}
\end{table}

To construct a comprehensive dataset tracing the evolution of Chinese characters from oracle bone script to contemporary forms, it is necessary to collect the written forms of the same character from different historical periods. This endeavor has been made feasible thanks to the diligent preservation of ancient texts by scholars throughout the centuries. Our work builds upon this legacy, creating the proposed \dataset{} by drawing upon classic works in the field of grammatology.

The evolution of Chinese characters can be divided into six distinctive periods according to their historical development, as illustrated in Figure~\ref{fig:intro-development}. These periods are the Oracle Bone Character (OBC) period, the Bronze Inscriptions (BI) period, the Seal Script (SS) period, the Spring and Autumn Characters (SAC) period, the Warring States Characters (WSC) period, and the Clerical Script (CS) period, collectively spanning nearly two thousand years. For each period, we have selected authoritative data sources for the corresponding Chinese character forms, guided by experts in oracle bone script studies. As shown in Table~\ref{table:data-source}, \dataset{} has gathered data from two sources: web-based databases and ancient script research books. Details of each source are introduced as follows:

\begin{figure}[t!]
    \centering
    \includegraphics[width=0.85\linewidth]{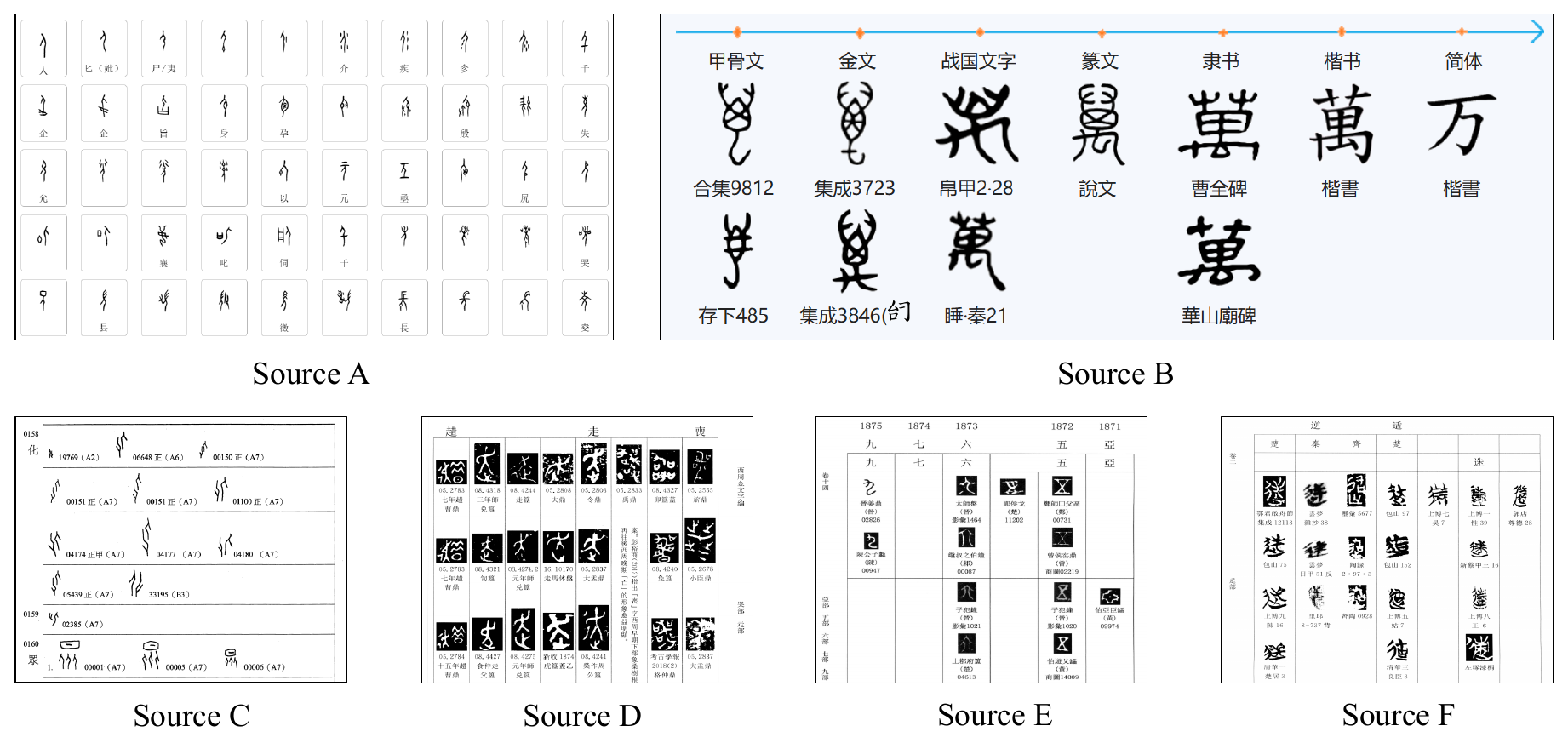}
    \caption{Samples from different data sources.}
    \label{fig:source-sample}
\end{figure}

\noindent\textbf{A. YinQiWenYuan} (\zh{殷契文渊})\footnote{\href{https://jgw.aynu.edu.cn/}{https://jgw.aynu.edu.cn/}} stands as one of the most extensive online repositories for oracle bone script, a key to understanding ancient Chinese writing system. Its collection comprises over a hundred varieties of oracle bone fragments, unearthed since the 19th century. These fragments have been meticulously cataloged and transcribed, featuring thousands of both deciphered and undeciphered oracle bone characters.

\noindent\textbf{B. GuoXueDaShi} (\zh{国学大师})\footnote{\href{https://www.guoxuedashi.net/zixing/yanbian/}{https://www.guoxuedashi.net/zixing/yanbian/}} has provided an online database for the evolution of Chinese characters, tracing their development from ancient pictographic forms to the modern standard simplified script.

\noindent\textbf{C. Oracle Bone Character Compilation} (\zh{甲骨文字编})\footnote{\href{https://books.google.com/books?id=4fuEwgEACAAJ}{https://books.google.com/books?id=4fuEwgEACAAJ}} is a systematically compiled dictionary of oracle bone characters. It transcribes inscriptions from oracle bones excavated up to the year 2010, encompassing a total of 4,378 individual characters, of which 1,682 can be interpreted.

\noindent\textbf{D. Compilation of Western Zhou Bronze Inscription} (\zh{西周金文字编})\footnote{\href{https://books.google.com/books?id=FUE6vwEACAAJ}{https://books.google.com/books?id=FUE6vwEACAAJ}} selects the most representative bronze inscriptions from the Western Zhou Dynasty, systematically and comprehensively compiling single characters from inscriptions found on bronze vessels of different periods unearthed from the Western Zhou era.

\noindent\textbf{E. Spring and Autumn Script Glyph Table} (\zh{春秋文字字形表})\footnote{\href{https://books.google.com/books?id=5J5sswEACAAJ}{https://books.google.com/books?id=5J5sswEACAAJ}} catalogs thousands of Chinese characters, recording their inscriptions as they appeared on bronze and stone artifacts from various feudal states during the Spring and Autumn period.

\noindent\textbf{F. Table of Glyphs for Warring States} (\zh{战国文字字形表})\footnote{\href{https://books.google.com/books?id=mU87tAEACAAJ}{https://books.google.com/books?id=mU87tAEACAAJ}} compiles an extensive assortment of inscriptions from a variety of artifacts, including bronzes, coins, pottery, and bamboo slips, striving to comprehensively showcase the full breadth of script styles prevalent during the Warring States period.

The data sources mentioned above collect both deciphered and undeciphered oracle bone characters. However, for the construction of an evolution dataset for Chinese characters, only the deciphered portions are needed. In Table~\ref{table:data-source}, the columns of \emph{\#Categories} and \emph{\#Images} show the number of individual character categories and images extracted from each source after filtering.

\subsection*{Digitalization}

\begin{figure}[t]
\centering
\includegraphics[width=\linewidth]{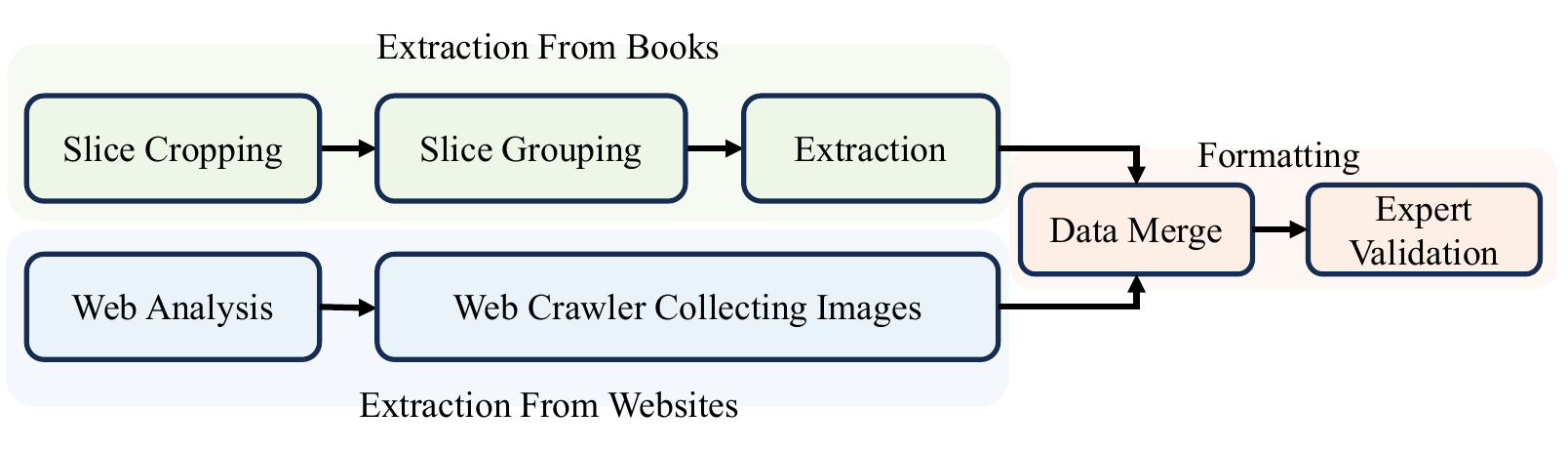}
\caption{Pipeline of digitalization of characters from different sources.}
\label{fig:pipeline}
\end{figure}

As illustrated in Figure~\ref{fig:source-sample}, building a dataset for the evolution of Chinese characters requires extracting text from a variety of data sources, each corresponding to different historical periods. These sources, while systematically compiled, often present a loosely organized structure. For instance, even though the ancient character compilations mentioned earlier are organized in dictionary order, the size and placement of the text images within them vary. This inconsistency makes direct extraction of the texts challenging, and manually scanning and cropping these data would be both laborious and costly. Therefore, to create a formatted dataset that supports computer-assisted research in Oracle bone script evolution, we have designed automated image extraction and categorization pipelines tailored to different sources (see Figure~\ref{fig:pipeline}).

\subsubsection*{Extraction from Books}

As illustrated in Souce C-F of Figure~\ref{fig:source-sample}, the source books typically record text in a table-like format, where each column or row, which we refer to as a \emph{slice}, contains several scanned images of ancient Chinese characters along with their corresponding source numbers. All images within a single slice correspond to the same modern Chinese character. However, each modern Chinese character may be associated with one or more slices from different sources. The goal of digitalization is to methodically extract and categorize the image patches in each slice, aligning them with their corresponding modern Chinese characters. To this end, as shown in Figure~\ref{fig:pipeline}, we have developed a three-step pipeline to automatically extract the samples from books, which comprises slice cropping, slice grouping, and extraction.

\begin{figure}[t]
\centering
    \subfigure[Original Book Page]{\includegraphics[width=0.33\linewidth]{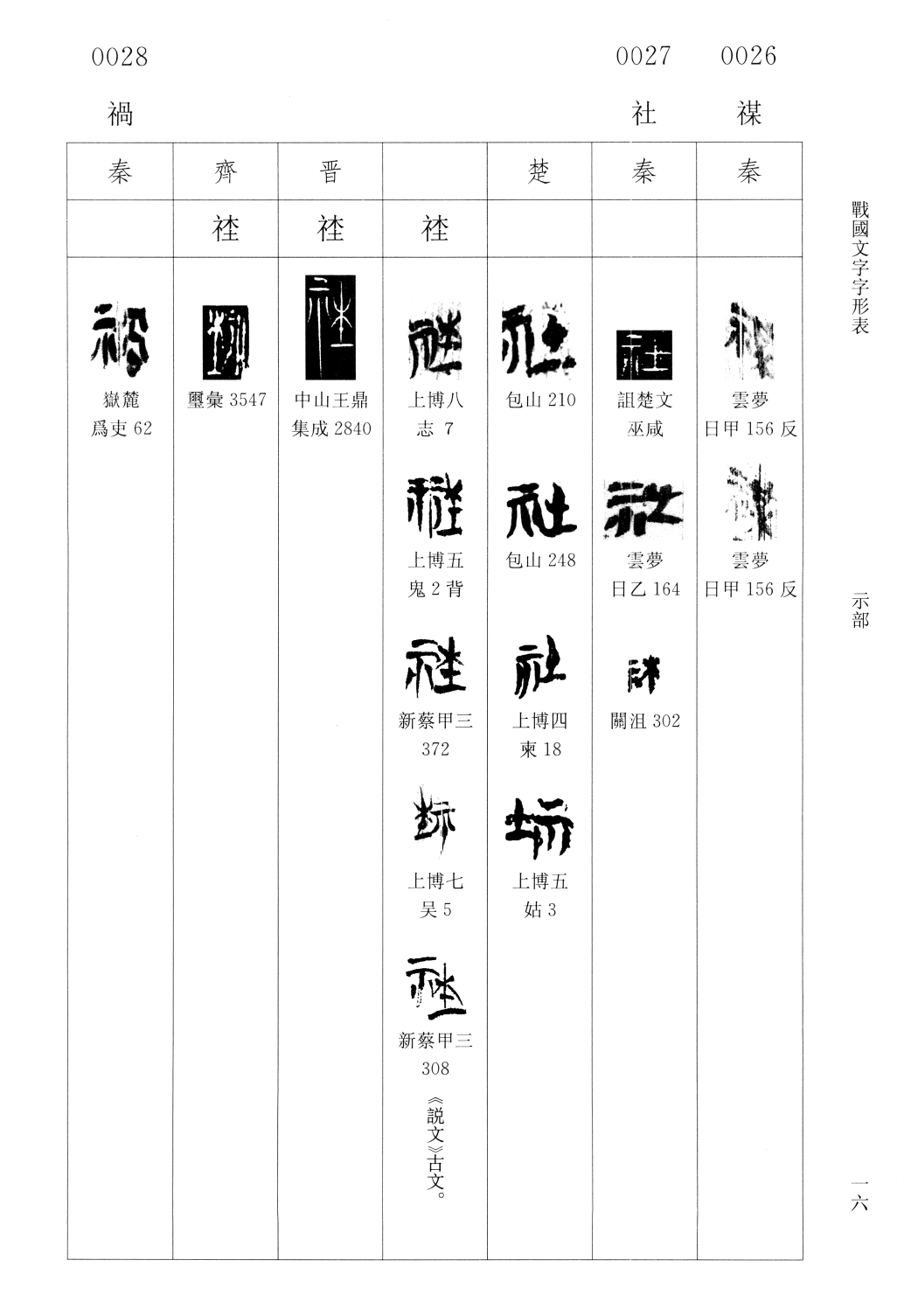}\label{fig:books-origin}}
    \subfigure[Slice Cropping]{\includegraphics[width=0.33\linewidth]{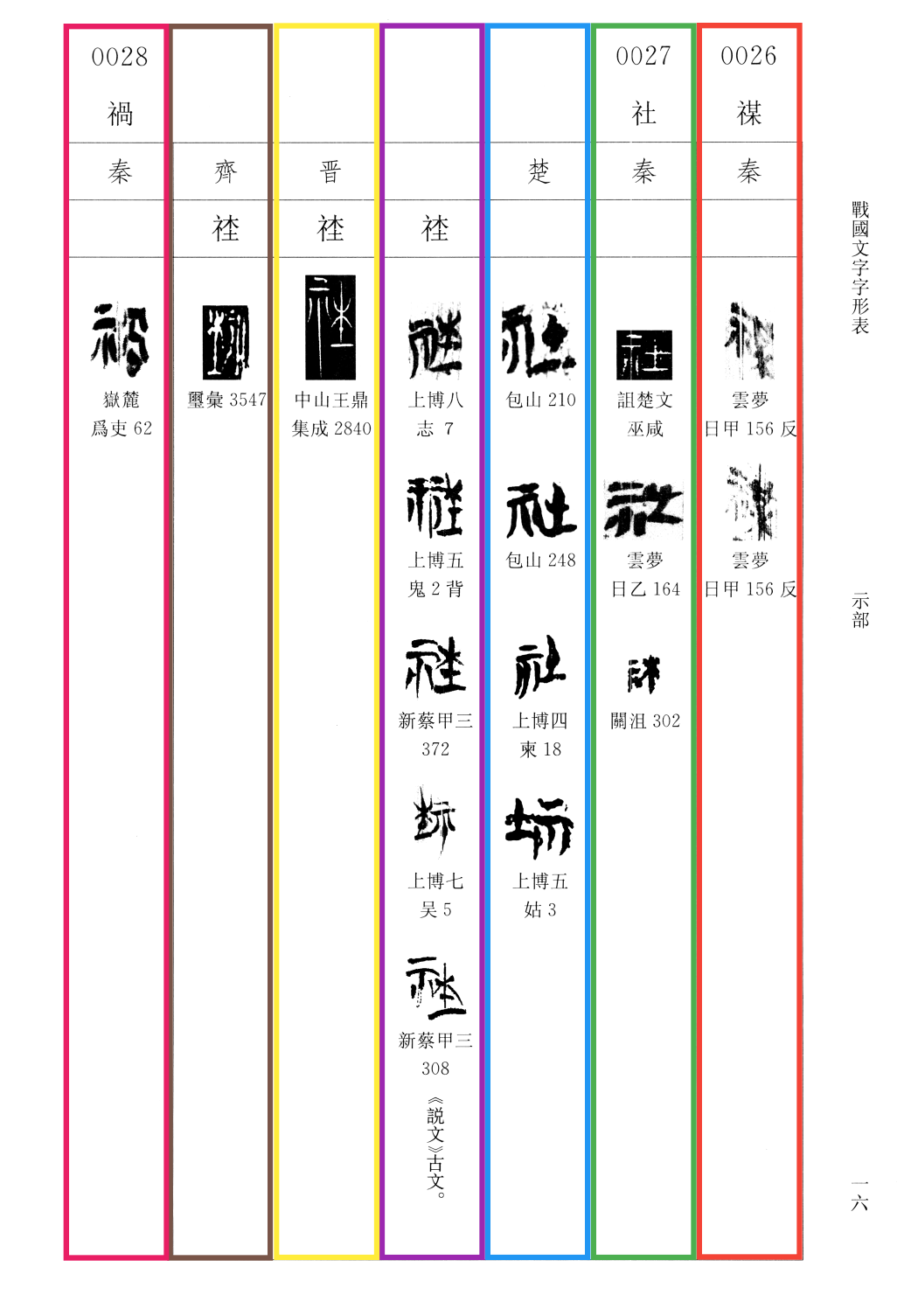}\label{fig:books-crop}}
    \subfigure[Slice Grouping]{\includegraphics[width=0.33\linewidth]{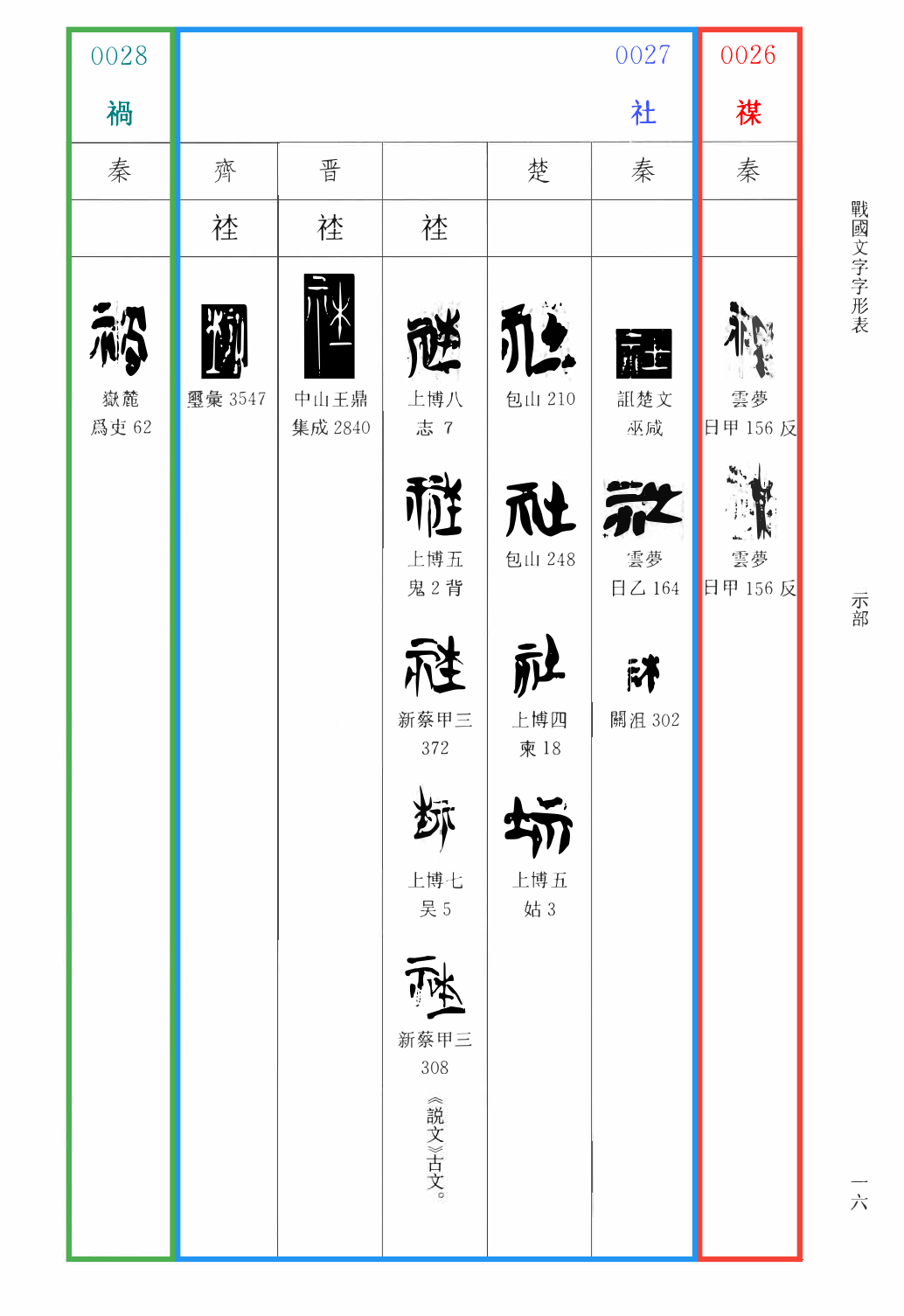}\label{fig:books-group}}
\caption{Procedure of cropping and grouping slices from book pages.}
\label{fig:books}
\end{figure}

\begin{figure}[t]
  \centering
  \includegraphics[width=0.9\linewidth]{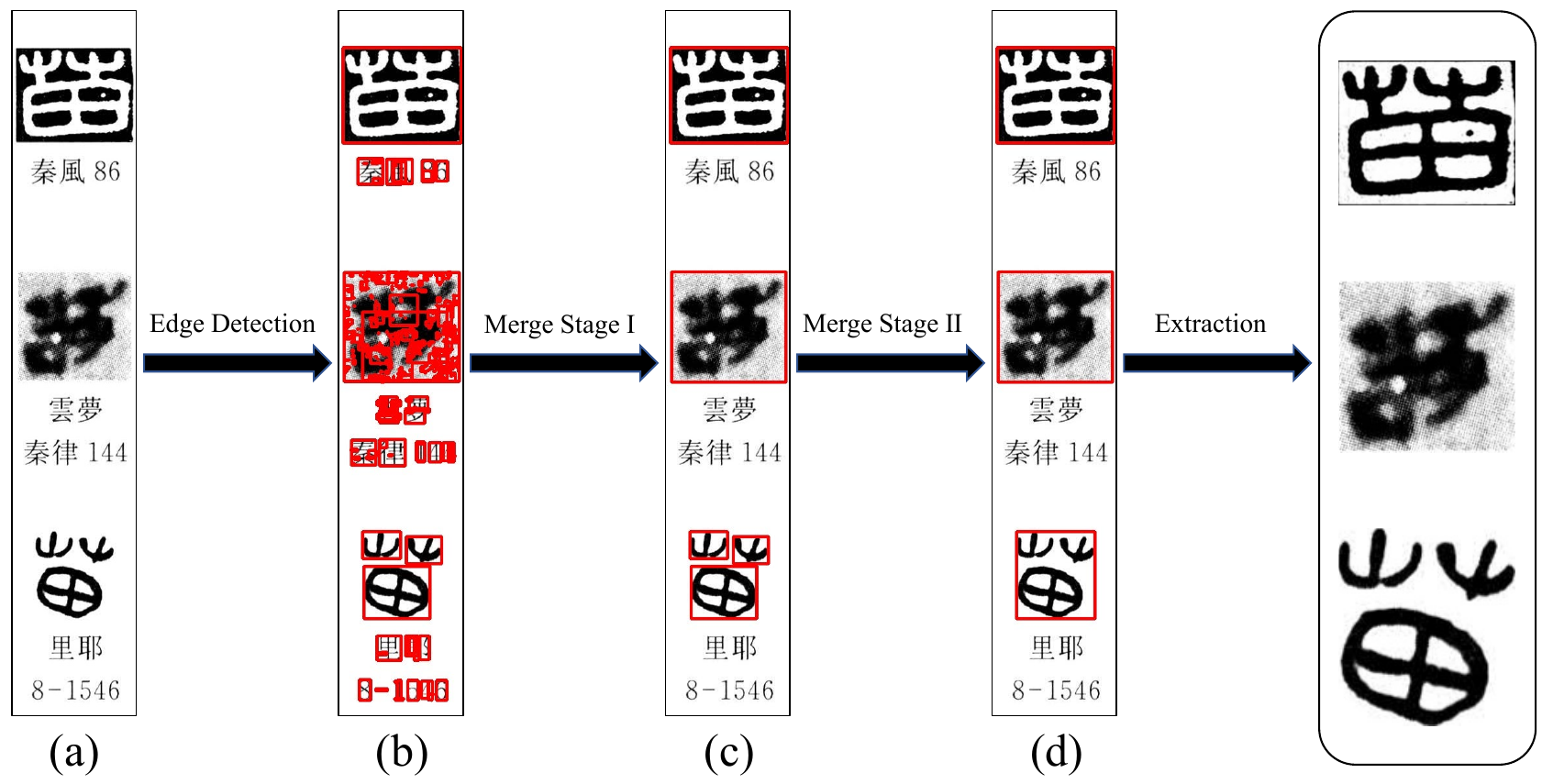}
  \caption{The pipeline of {\bfseries IMNNB}. (a) is the original slice cropped from a book page. (b) shows the bounding boxes generated by edge detection algorithms (c) and (d) illustrate the slice images after Merge Stage I and Merge Stage II, respectively.}
  \label{fig:IMNNB}
\end{figure}

\begin{enumerate}
    \item\textbf{Slice Cropping} is developed to cut out slices from each page, adhering to the book's layout. Taking the \textit{Table of Glyphs for Warring States} as an example, Figure~\ref{fig:books-origin} shows a page from this book. It can be observed that this book is organized in a multi-column table-like layout, which is used to present the variations of ancient Chinese characters across different states during the Warring States period. The table is organized meticulously: the first row names the feudal kingdom associated with the character, the second shows ancient variants of the character, and the third row presents scanned images of the character alongside their corresponding source codes. At the top of the page, outside the table, two additional lines of \emph{header information} are provided: the category number of the character in this book, and the corresponding standard modern Chinese character. The book's layout follows a right-to-left reading order, indicating that when there is a space above a slice without text codes, it falls under the first code category that appears to its right. Having grasped the layout information of the book, the first step we need to take is to extract the \emph{slices} of each page as shown in Figure~\ref{fig:books-crop}. Thanks to the book being formatted in a tabular form, the most straightforward method is to employ edge detection algorithms to identify the table borders. Specifically, we employed the OpenCV library to detect black edges in the images and then standardized these into rectangular boxes. Due to the presence of horizontal borders, the tables may be divided into multiple rows of boxes. Therefore, we further merge and expand these rectangles vertically, thus dividing a page of the book into several long slices as illustrated in Figure~\ref{fig:books-crop}.
    \item\textbf{Slice Grouping} is responsible for assigning category labels to each slice. After the Slice Cropping step, the entire book has been divided into a multitude of slices. However, these slices still lack labels, meaning it is unknown which modern Chinese character corresponds to the ancient character images in these slices. Therefore, to automate the grouping of slices of the same category and label them, it is necessary to recognize the \textit{header information} of each slice. Specifically, we first sort the slices in the reading order from right to left, and then crop the top area of each slice to feed it into an Optical Character Recognition (OCR) system. This OCR system performs two stages: text detection and text recognition. The purpose of text detection is to determine whether the head of the current slice is empty. If it is not empty, this indicates that the current and all subsequent slices until the next non-empty head is detected, belong to the same category. The text recognition aims to identify the modern standard Chinese characters present in the header, which are then used as category labels assigned to each corresponding slice. After this procedure, each slice can then be automatically grouped and labeled (see Figure~\ref{fig:books-group}).
    \item\textbf{Extraction} is the final step, used to retrieve the image patches of the oracle bone character rubbings from the annotated slices. As shown in Figure~\ref{fig:IMNNB}(a), each slice contains one or more scanned images of ancient characters. These image patches vary in size and lack distinct borders for separation, blending with the provenance annotations of modern Chinese text below. This complexity renders traditional edge detection methods, like those used in the slice cropping stage, ineffective. To address this, we have developed Iterative Merging of Nearest Neighbor Boxes (IMNNB) for separating these image patches of ancient texts from backgrounds. As illustrated in Figure~\ref{fig:IMNNB}, IMNNB consists of three steps: Edge Detection, Merge Stage I, and Merge Stage II.
    \begin{enumerate}
        \item\textbf{Edge Detection} is initially employed to generate edge bounding boxes, which serve to distinguish between the foreground and the blank background in each slice. As illustrated in Figure~\ref{fig:IMNNB}(b), although this step can effectively separate the foreground texts from the blank areas of the book page, it also inadvertently includes unwanted regions, such as image noise and provenance annotations.
        \item\textbf{Merge Stage I} is designed to filter out noise. As seen in Figure~\ref{fig:IMNNB}(b), there is a significant overlap among the bounding boxes generated by the edge detection algorithm. These edges, created by internal details and noise of each item in the slice, are unnecessary for extracting the scanned image patches of ancient text. Therefore, we employ an iterative merging method to merge these overlapping bounding boxes, thereby reducing their quantity. Specifically, any two bounding boxes with overlapping areas are merged into their combined area. Ultimately, the entire slice will only contain independent bounding boxes with a zero intersection-over-union ratio to each other. Furthermore, considering the unique position and smaller font of provenance labels under each scanned image, they can be directly filtered out using a predetermined threshold. In practice, any bounding box smaller than 2,000 pixels is excluded. As demonstrated in Figure~\ref{fig:IMNNB}(c), this step effectively removes irrelevant noise, yielding relatively complete bounding areas for the ancient text images.
        \item\textbf{Merge Stage II} is used for combining the radicals and components of ancient characters. As the bottom character shown in Figure~\ref{fig:IMNNB}(c), when the components of an ancient character are independent and separated by considerable gaps, it results in the appearance of multiple bounding boxes within a single character. To address this issue, it is essential to further merge these bounding boxes into a single cohesive text unit. Assume a slice retains $N$ bounding boxes after passing through Merge Stage I. Let $B$ represent the set of bounding boxes, with each box denoted as $b_{i}$, where $b_{i} \in B$ and $i$ ranges from $1$ to $N$. For $\forall b_{i}, b_{j} \in B, \text{ where } i \neq j$, within their center coordinates at $\left(x_{i}, y_{i}\right)$ and $\left(x_{j}, y_{j}\right)$, respectively. We then calculate the vertical distance between the centers of the box $b_{i}$ and box $b_{j}$, denoted as $\left| y_{i} - y_{j} \right|$. All bounding boxes with distances less than a threshold value will be merged, specifically $\left| y_{i} - y_{j} \right| < \tau$, where practically, $\tau$ is set to 150 pixels. Following this step, as shown in Figure~\ref{fig:IMNNB}(d), it can successfully capture the entire bounding box of the scanned image of the ancient character, allowing us to extract the corresponding patch from each slice.
    \end{enumerate}
     Algorithm~\ref{alg:IMNNB} provides a detailed pseudocode of the entire extraction process.
\end{enumerate}

It is noteworthy that, although there are subtle differences in the layouts of various books used for building \dataset{}, such as reading order and font size, they generally follow a similar tabular format (as shown in Figure~\ref{fig:source-sample} source C-F). Therefore, the process we proposed can be seamlessly applied to these books for extracting images of ancient texts.

\begin{algorithm}[b!]
  \SetKwInOut{Input}{Input}\SetKwInOut{Output}{Output}
  \Input{A vertical slice {\bfseries\emph{S}}}
  \Output{Boxes for ancient Chinese characters {\bfseries\emph{B}}}
  \KwData{Center coordinate of {\bfseries\emph{B[i]}} ($x_{i}$, $y_{i}$)}
  {\bfseries\emph{B[i]}} $\leftarrow$ Edge detection({\bfseries\emph{S}}) \tcp*[r]{i = 1,2,...,n}\
  \tcp{Merge Stage \uppercase\expandafter{\romannumeral1}}\
  \While{exist two intersections boxes {\bfseries B[i]} and {\bfseries B[j]} in the {\bfseries B}}{
    \emph{B}\textsubscript{m} $\leftarrow$ Merge operation({\bfseries \emph{B[i]}}, {\bfseries \emph{B[j]}})\;
    Delete {\bfseries \emph{B[i]}} and {\bfseries \emph{B[j]}}\;
    {\bfseries\emph{B}}.append(\emph{B}\textsubscript{m})\;
  }\
  \tcp{Remove boxes that are too small}\
  \For{{\bfseries B[i]} in {\bfseries B}}{
  \If{the area of {\bfseries B[i]} < 2000}{
  Delete {\bfseries \emph{B[i]}}
  }
  }\
  \tcp{Merge Stage \uppercase\expandafter{\romannumeral2}}\
  \While{exist two vertical coordinate $\lvert y_{i}-y_{i} \rvert$ < 150}{
    \emph{B}\textsubscript{m} $\leftarrow$ Merge operation({\bfseries \emph{B[i]}}, {\bfseries \emph{B[j]}})\;
    Delete {\bfseries \emph{B[i]}} and {\bfseries \emph{B[j]}}\;
    {\bfseries\emph{B}}.append(\emph{B}\textsubscript{m})\;
  }
  \Return{{\bfseries B}}\;
\caption{{\bfseries Iterative Merging of Nearest Neighbor Boxes (IMNNB)}}
\label{alg:IMNNB}
\end{algorithm}

\begin{table}[b!]
\centering
\begin{tabular}{lcc}
\hline
\multicolumn{1}{c|}{Era} & \multicolumn{1}{c|}{\#Categories} & \#Images \\ \hline
\multicolumn{3}{c}{Data from Books} \\ \hline
\multicolumn{1}{l|}{Oracle Bone Characters} & \multicolumn{1}{c|}{1,590} & 27,276 \\
\multicolumn{1}{l|}{Bronze Inscriptions} & \multicolumn{1}{c|}{2,040} & 21,681 \\
\multicolumn{1}{l|}{Spring \& Autumn Characters} & \multicolumn{1}{c|}{1,905} & 9,131 \\
\multicolumn{1}{l|}{Warring States Characters} & \multicolumn{1}{c|}{6,658} & 32,794 \\
\multicolumn{1}{l|}{\textcolor{gray}{Subtotal}} & \multicolumn{1}{c|}{\textcolor{gray}{8,376}} & \textcolor{gray}{90,882}              \\ \hline
\multicolumn{3}{c}{Data from Websites} \\ \hline
\multicolumn{1}{l|}{Oracle Bone Characters} & \multicolumn{1}{c|}{1,487} & 48,405 \\
\multicolumn{1}{l|}{Bronze Inscriptions} & \multicolumn{1}{c|}{2,729} & 25,633 \\
\multicolumn{1}{l|}{Seal Script} & \multicolumn{1}{c|}{9,147}                    & 13,434 \\
\multicolumn{1}{l|}{Warring States Characters} & \multicolumn{1}{c|}{3,111} & 47,248 \\
\multicolumn{1}{l|}{Clerical Script} & \multicolumn{1}{c|}{2,890} & 3,568 \\
\multicolumn{1}{l|}{\textcolor{gray}{Subtotal}} & \multicolumn{1}{c|}{\textcolor{gray}{10,207}} & \textcolor{gray}{138,288} \\ \hline
\multicolumn{1}{l|}{Total} & \multicolumn{1}{c|}{13,714} & 229,170 \\\hline
\end{tabular}
\caption{Statistics of category and image counts from different sources.}
\label{tab:data-summary}
\end{table}

\subsubsection*{Extraction from Websites}

For website data, we select two large-scale ancient script repositories, \emph{i.e.}, YinQiWenYuan~\cite{YinQiWenYuan} and GuoXueDaShi~\cite{GuoXueDaShi}, to employ web crawlers for collecting corresponding images and labels. Fortunately, these websites already offer ancient script images that are cropped and aligned, sparing us the additional steps of detection and segmenting that are necessary when extracting text from books.

\subsubsection*{Formatting}

Due to the use of samples from various sources, including different websites and books, in the dataset construction process, the formats of the original data are not uniform. To ensure consistency in the samples of the final product, it is necessary to standardize the formatting of the extracted data. This primarily includes the following aspects:

\textbf{Background Normalization:} Varying scanning methods for ancient text images in different sources may result in images with either black text on a white background or white text on a black background. To maintain a consistent data distribution, we have normalized all images in the \dataset{} to have a white background with black text.

\textbf{Category Unification:} Due to the mixing of Simplified and Traditional Chinese characters in the annotations of ancient text from different sources, images of the same category may be divided into two separate groups: one for Simplified and the other for Traditional characters. To eliminate these redundant categories, we merged them by referring to a conversion table between Simplified and Traditional Chinese characters.

\textbf{Naming:} For the convenience of using the dataset and retrieving data, we have uniformly named all the images following this format Source\_Era\_ID. Here \emph{Source} indicates whether the image originates from the Internet or a book, while \emph{Era} refers to the period of the text, which could be one of the following: Oracle Bone Character, Bronze Inscriptions, Seal Script, Spring \& Autumn Characters, Warring States Characters, and Clerical Script. \emph{ID} represents the sample number.

Despite the majority of steps in constructing the dataset being completed by an automated pipeline, we further conducted a thorough manual review of the extracted images and labels to ensure the quality of the final dataset. Specifically, we compared the images extracted from books with their original manuscripts and double-checked the labels for each image. After eliminating or correcting images of low quality and incorrect annotations, we submitted the dataset to external experts in Oracle bone script research, entrusting them to assess the samples and labels within the dataset for quality evaluation. After completing all the above steps, we have obtained the final product, which we have named the \textbf{EV}olution of \textbf{O}racle \textbf{B}one \textbf{C}haracters (\textbf{EVOBC}).

\section*{Data Records}

This paper introduces the \dataset{}, which ultimately comprises 229,170 images across 13,714 categories (please refer to Table~\ref{tab:data-summary} for detailed statistical data). The dataset is organized into 9 folders based on the source and period, namely Book\_Oracle, Book\_Bronze, Book\_SprAut, Book\_War, Website\_Oracle, Website\_Bronze, Website\_War, Website\_Seal, and Website\_Clerical. Each folder contains several subfolders named after categories, housing scanned images of ancient Chinese characters sourced from various origins. Additionally, we provide annotation files in JSON format containing metadata such as image paths, categories, image sizes, etc. Table~\ref{tab:samples} presents some examples from the \dataset{}. For each character, we have collected its representations from up to six major historical periods. The images for each period are composed of samples from various sources and may include up to hundreds of different images from diverse origins.

\begin{table}[h!]
\renewcommand\arraystretch{1.5}
    \centering
    \begin{tabular}{|c|c|c|c|c|c|c|c|c|}
    \hline
         & &\multicolumn{6}{c|}{Scanned Images of Texts from Different Eras} & \\\cline{3-8}
        Standard Modern Character & Source & \textbf{OBC} & \textbf{BI} & \textbf{SS} & \textbf{SAC} & \textbf{WSC} & \textbf{CS} & \#Samples \\\hline
         \multirow{2}{*}{\includegraphics[width=1cm]{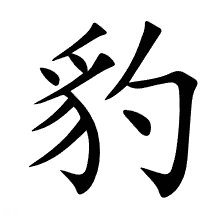}} & Book & \raisebox{-0.1\height}{\includegraphics[height=1cm]{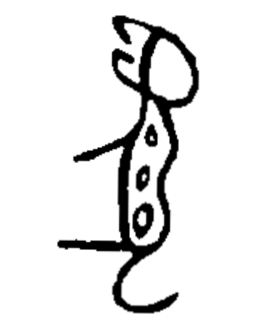}} & \includegraphics[height=1cm]{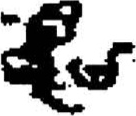} &  &  & \includegraphics[height=1cm]{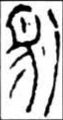} & & \multirow{2}{*}{90} \\\cline{2-8}
         & Website & \raisebox{-0.1\height}{\includegraphics[height=1cm]{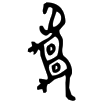}} & \includegraphics[height=1cm]{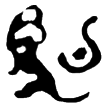} &  \includegraphics[height=1cm]{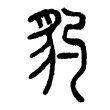} &  & \includegraphics[height=1cm]{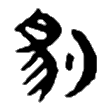} & \includegraphics[height=1cm]{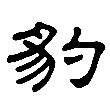} & \\\hline
          \multirow{2}{*}{\includegraphics[width=1cm]{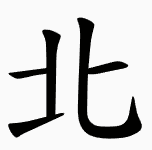}} & Book & \raisebox{-0.1\height}{\includegraphics[height=1cm]{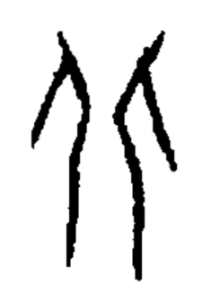}} & \raisebox{-0.1\height}{\includegraphics[height=1cm]{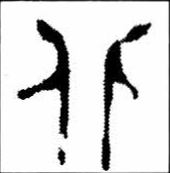}} & & \raisebox{-0.1\height}{\includegraphics[height=1cm]{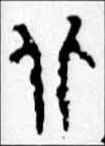}} & \includegraphics[height=1cm]{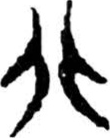} & & \multirow{2}{*}{230} \\\cline{2-8}
         & Website & \raisebox{-0.1\height}{\includegraphics[height=1cm]{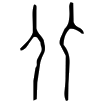}} & \includegraphics[height=1cm]{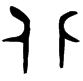} & \includegraphics[height=1cm]{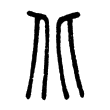} & & \includegraphics[height=1cm]{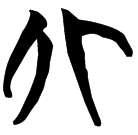} & \includegraphics[height=1cm]{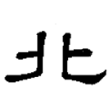} & \\\hline
         \multirow{2}{*}{\includegraphics[width=1cm]{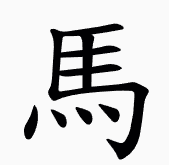}} & Book & \raisebox{-0.1\height}{\includegraphics[height=1cm]{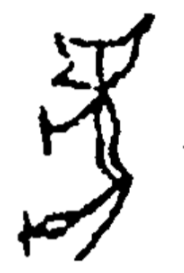}} & \raisebox{-0.1\height}{\includegraphics[height=1cm]{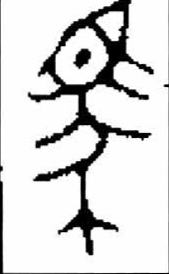}} &  & \raisebox{-0.1\height}{\includegraphics[height=1cm]{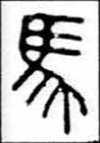}} & \includegraphics[height=1cm]{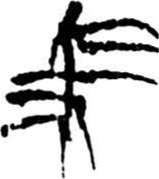} & & \multirow{2}{*}{520} \\\cline{2-8}
         & Website & \raisebox{-0.1\height}{\includegraphics[height=1cm]{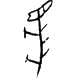}} & \includegraphics[height=1cm]{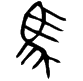} & \includegraphics[height=1cm]{figures/sample-table/bao/seal.png} & & \includegraphics[height=1cm]{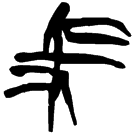} & \includegraphics[height=1cm]{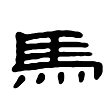} & \\\hline
         \multirow{2}{*}{\includegraphics[width=1cm]{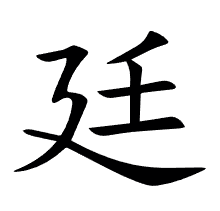}} & Book & & \includegraphics[height=1cm]{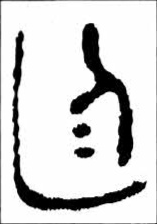} & & \raisebox{-0.1\height}{\includegraphics[height=1cm]{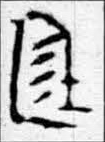}} & \includegraphics[height=1cm]{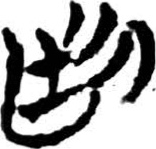} & & \multirow{2}{*}{162} \\\cline{2-8}
         & Website &  & \raisebox{-0.1\height}{\includegraphics[height=1cm]{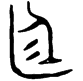}} & \raisebox{-0.1\height}{\includegraphics[height=1cm]{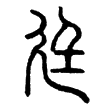}} & & \raisebox{-0.1\height}{\includegraphics[height=1cm]{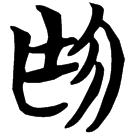}} & \raisebox{-0.1\height}{\includegraphics[height=1cm]{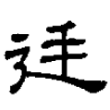}} & \\
         \hline            
    \end{tabular}
    \caption{Samples showing the evolution of different Chinese characters in \dataset{}.}
    \label{tab:samples}
\end{table}

\section*{Technical Validation}

To demonstrate the utility of the proposed EVOBC dataset in computer-assisted research of oracle bone inscriptions, we devised two types of technical validation tasks from the perspective of artificial intelligence, particularly computer vision. These tasks are Image Classification and the Oracle Bone Character Deciphering Simulation.

\subsection*{Image Classification}

\begin{figure}[t!]
\centering
    \subfigure[ResNet-101]{\includegraphics[width=0.45\linewidth]{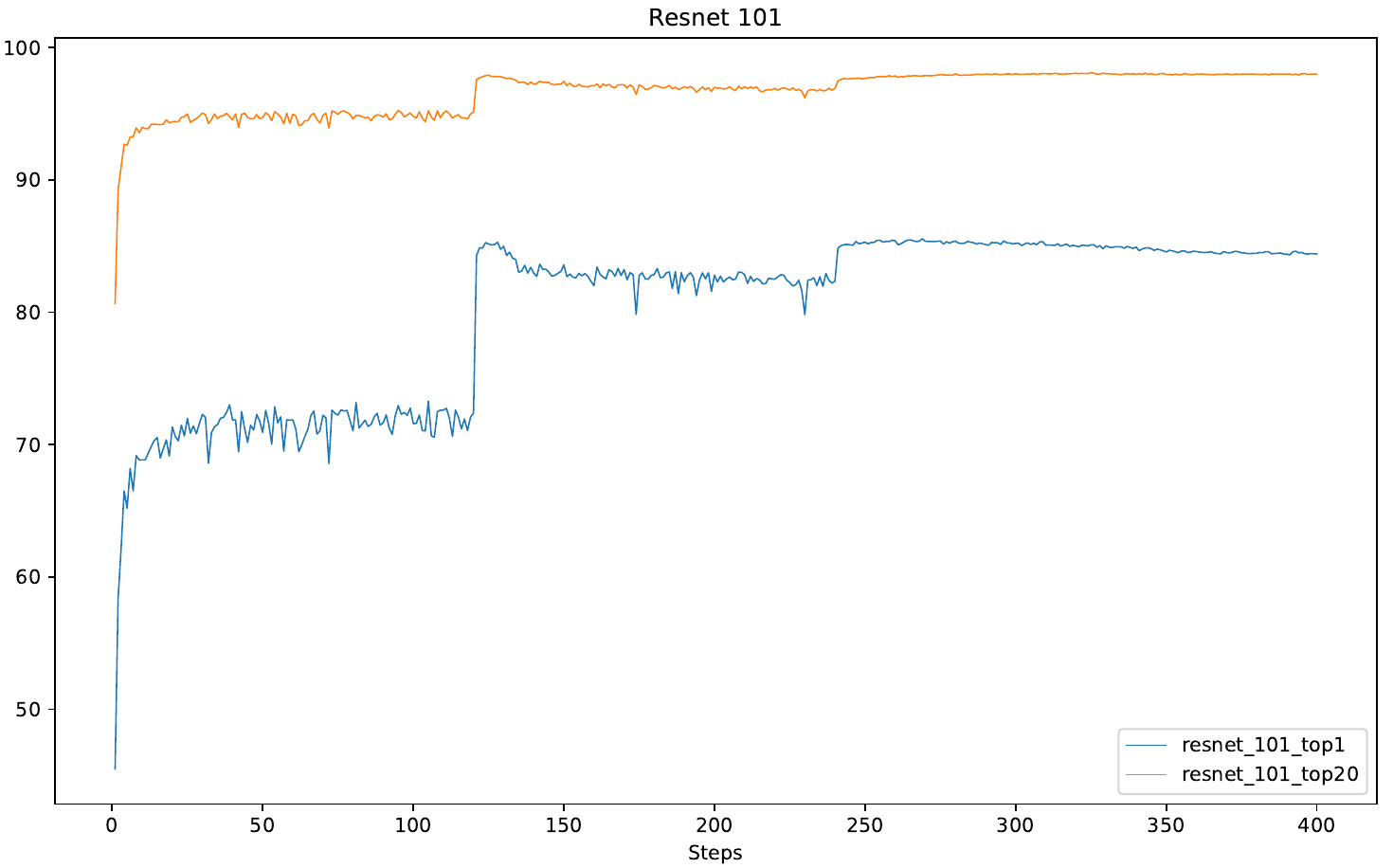}}
    \subfigure[Swin-Transformer v2]{\includegraphics[width=0.45\linewidth]{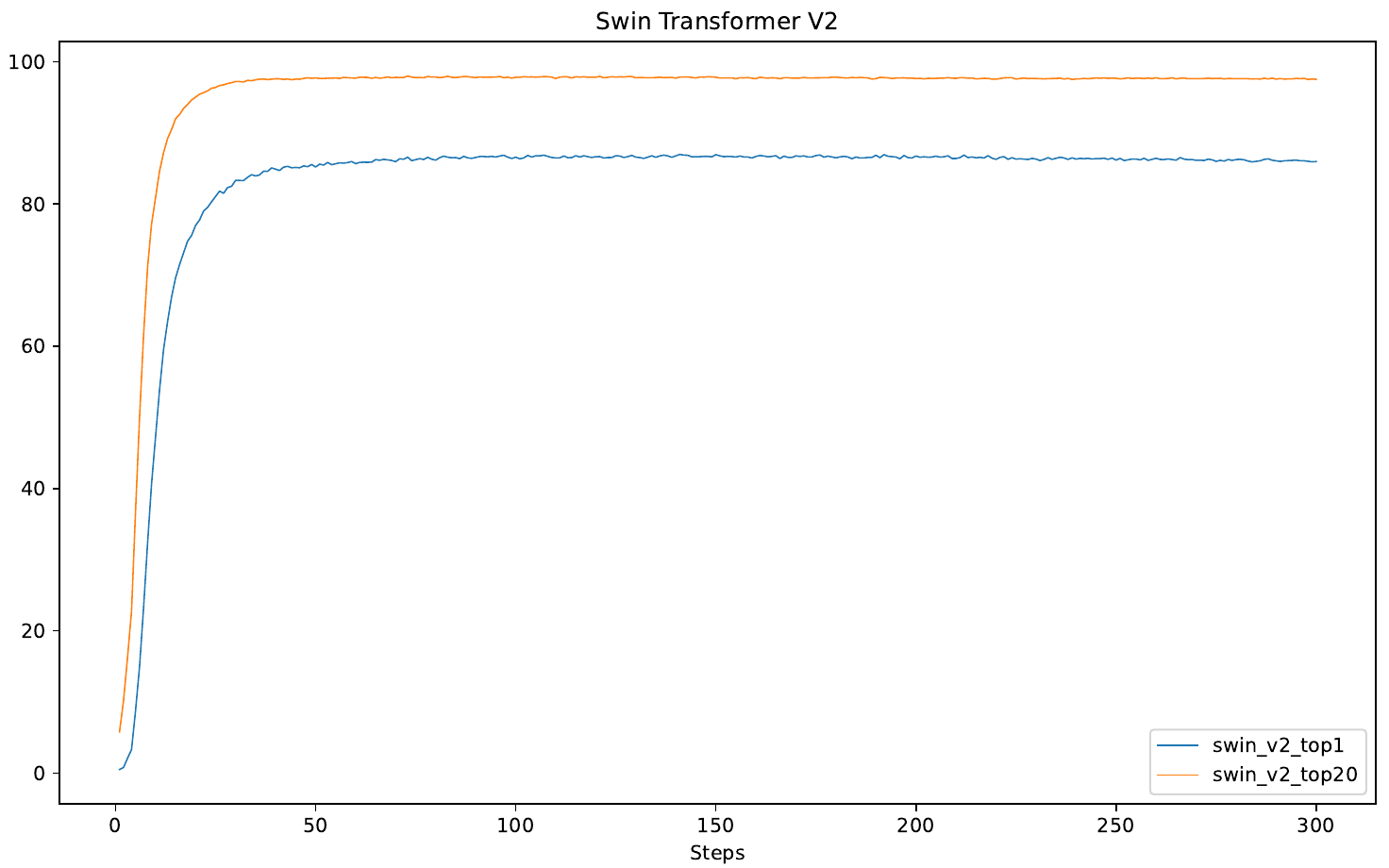}}
\caption{Top-1 and Top-20 accuracy of ResNet-101 and Swin-Transformer v2 on the \dataset{} dataset.}
\label{fig:classification-res}
\end{figure}

Image classification is one of the fundamental tasks in the field of computer vision. It involves training a model with a certain number of images, enabling the model to automatically categorize new coming images. The \dataset{} dataset provides unique category labels for each ancient script image, making it highly suitable for validating the quality of the dataset through the image classification task. The accuracy of the final classification model serves as a validation measure; a low accuracy would indicate a large number of erroneous or unreasonable annotations in the dataset, whereas a small classification error would signify a high-quality dataset. Specifically, we divided the \dataset{} dataset into training and validation sets in a 9:1 ratio, using them to train and test two widely used classification models: ResNet-101~\cite{he2016deep} and Swin Transformer v2~\cite{liu2022swin}. The results, including the Top-1 and Top-20 classification accuracy rates at various training steps, were displayed in Figure~\ref{fig:classification-res}, while Table~\ref{tab:classification-res} showcased the best performance. Despite the challenging nature of classifying oracle bone script, both models achieved commendable performance, with Top-1 scores of 85.56\% for ResNet-101 and 86.66\% for Swin Transformer v2, demonstrating the high quality of EVOBC dataset annotations and its potential value in future AI-assisted oracle bone script research.

\begin{table}[t!]
    \centering
    \begin{tabular}{c|c|c} \hline 
         Model & Top-1 Acc & Top-20 Acc \\ \hline 
         ResNet-101 & 85.56\% & 97.85\% \\ 
         Swin Transformer v2 &  86.66\% & 97.93\% \\ \hline
    \end{tabular}
    \caption{Quantitative results of classification task on the validation set of \dataset{}.}
    \label{tab:classification-res}
\end{table}

\subsection*{Oracle Bone Character Deciphering Simulation}

One of our objectives in proposing the \dataset{} dataset is to aid future development in AI-assisted research of oracle bone script. By leveraging AI models, we aim to uncover the evolutionary patterns of ancient Chinese characters and assist in deciphering oracle bone inscriptions whose meanings are currently unclear. To this end, we have introduced a new task, Oracle Bone Character Deciphering Simulation, to test this possibility on the \dataset{} dataset. Given the uncharted territory of this research field, we have suggested two preliminary, yet potentially effective, approaches as baselines for this task: one based on \textit{Image Classification} and the other on \textit{Image Generation}.

\begin{figure}[t!]
\centering
    \subfigure[Quantitative Results of ResNet-101]{\includegraphics[height=4.5cm]{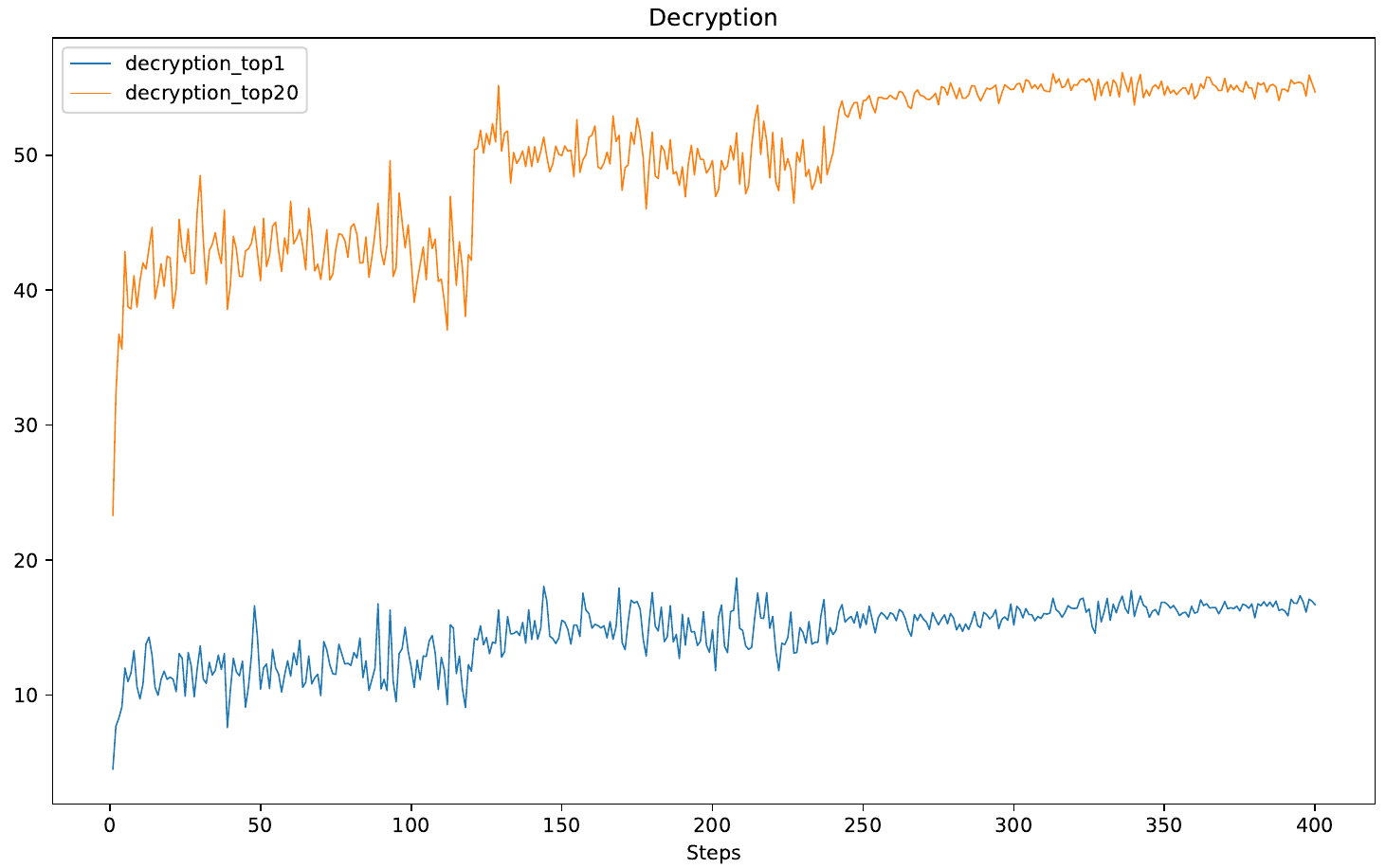}\label{fig:quantitative-resnet101}}
    \subfigure[Qualitative Results of Conditioned Diffusion Model]{\includegraphics[height=4.5cm]{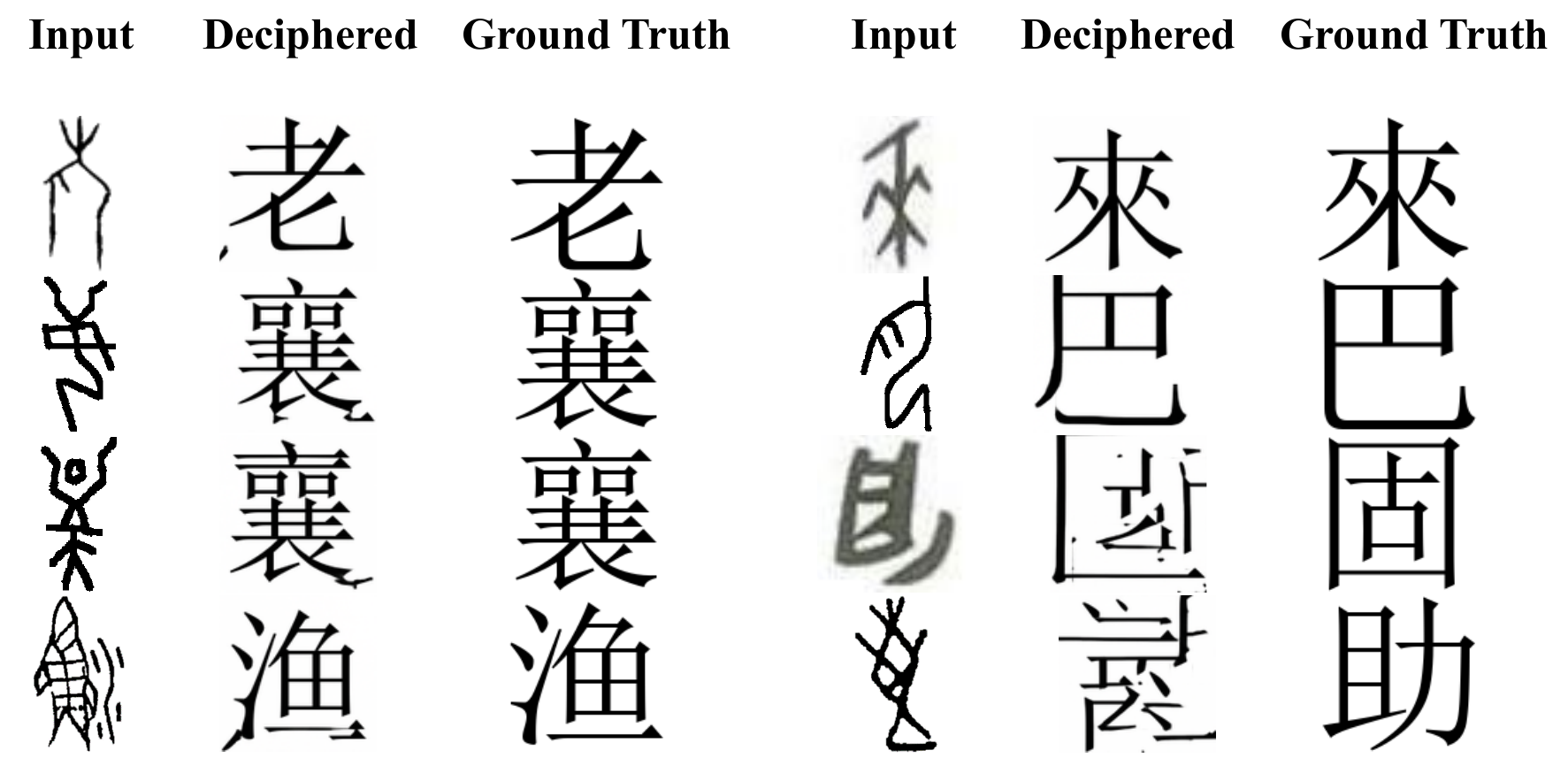}\label{fig:qualitative-diffusion}}
\caption{Results of the Oracle Character Deciphering Simulation task.}
\end{figure}

\textbf{Image Classification.} In the last section, the effectiveness of classification models on the \dataset{} dataset has already been demonstrated. Thus, we also attempted to adapt the classifier to deciphering oracle bone characters. Specifically, we trained the classifier using oracle bone inscriptions and their evolved counterparts from other eras as training samples. This approach enables the classifier to learn the associations and morphological transformations between the oracle bone script and texts from other periods. During the testing phase, the model could leverage the acquired patterns to correlate them with the most analogous characters from known texts of other eras, thereby achieving a description effect. Figure~\ref{fig:quantitative-resnet101} illustrates the accuracy of oracle bone script deciphering using ResNet-101 as the classifier on the \dataset{} dataset, achieving top-1 and top-20 accuracies of 16.7\% and 55.8\%, respectively. This not only showcases the challenging nature of deciphering oracle bone script but also attests to the potential value of the \dataset{} for future research.

\textbf{Image Generation.} Considering the rapid development and significant success of image generation algorithms recently, we have attempted to decipher oracle bone script from the perspective of image generation. To this end, we trained a conditional diffusion model~\cite{choi2021ilvr} on \dataset{} dataset. Specifically, we use original oracle bone script images as input conditions, and images of translated texts from other eras as the generation targets. In the testing phase, by inputting undeciphered oracle bone script images into the model, it can generate text images of the corresponding era. By comparing these generated images with known images, the deciphering process is completed. Figure~\ref{fig:qualitative-diffusion} shows the qualitative results of this model, where the \textit{Input} column represents the original oracle bone script images fed into the model, the \textit{Deciphered} column shows the text images generated by the conditioned diffusion model, and the \textit{Ground Truth} column is the corresponding real text labels. As we can see, this model has already demonstrated certain capabilities in deciphering the oracle bone script.

In summary, we conducted technical validation on our proposed \dataset{} for both image classification and oracle bone character deciphering simulation tasks, achieving highly encouraging results. This not only attests to the high-quality annotations of the \dataset{} but also establishes a baseline for future related research/ We hope that it will shed light on the path for future AI-assisted oracle bone script studies and serve as a cornerstone in this field.

\section*{Usage Notes}

To ensure the reproducibility of our work, we will release all the scripts and codes employed in building the \dataset{} as well as the final product. This will include, but not be limited to, data processing scripts for image denoising and alignment, tools for extracting and automatically annotating ancient Chinese scanned images from book sources, as well as the training and testing scripts used in our technical validation experiments. For more information, please see (\href {https://github.com/RomanticGodVAN/character-Evolution-Dataset.git}{https://github.com/RomanticGodVAN/character-Evolution-Dataset.git}).

\section*{Code availability}

Code for building \dataset{} will be made available at \href {https://github.com/RomanticGodVAN/character-Evolution-Dataset.git}{https://github.com/RomanticGodVAN/character-Evolution-Dataset.git}. We use the MMPretrain~\cite{2023mmpretrain} toolbox to conduct the technical validation experiments.

\bibliography{ref}

\section*{Acknowledgements} 

The authors thank the Key Laboratory of Oracle Bone Script Information Processing, Ministry of Education, Anyang Normal University for providing ancient text data sources and review of the dataset construction. 

\section*{Author Contributions Statement}

Haisu Guan conceived the technical validation experiment(s), Haisu Guan and Jinpeng Wan built the dataset from books and co-write this paper, and Pengjie Wang and Kaile Zhang built the dataset from the website. Jinpeng Wan completes the label identification and sorting of the dataset, Yuliang Liu guides the entire project, and Xiang Bai provides laboratory resources. All authors reviewed the manuscript. 

\section*{Competing Interests} 

The corresponding author is responsible for providing a \href{https://www.nature.com/sdata/policies/editorial-and-publishing-policies#competing}{competing interests statement} on behalf of all authors of the paper. This statement must be included in the submitted article file.

\end{document}